\pdfoutput=1

\documentclass[11pt]{article}

\usepackage[]{EMNLP2023} 

\usepackage{times}
\usepackage{latexsym}

\usepackage[T1]{fontenc}

\usepackage[utf8]{inputenc}

\usepackage{microtype}

\usepackage{inconsolata}

\usepackage{graphicx}
\usepackage{multirow}
\definecolor{airforceblue}{rgb}{0.36, 0.54, 0.66}
\definecolor{bazaar}{rgb}{0.6, 0.47, 0.48}
\definecolor{bulgarianrose}{rgb}{0.28, 0.02, 0.03}
\definecolor{cadet}{rgb}{0.33, 0.41, 0.47}
\usepackage{amsmath} 
\usepackage{soul}
\usepackage{adjustbox}
\usepackage{enumitem}
\usepackage{xcolor, soul}
\definecolor{shadecolor}{rgb}{0.9,0.9,0.9}
\sethlcolor{shadecolor}
\usepackage{array}
\usepackage{arydshln}

\NewDocumentCommand{\heng}
{ mO{} }{\textcolor{red}{\textsuperscript{\textit{Heng}}\textsf{\textbf{\small[#1]}}}}

\usepackage{makecell}

\title{\textsc{NormSage}: Multi-Lingual Multi-Cultural Norm Discovery \\ from Conversations On-the-Fly}

\author{\textsuperscript{$\spadesuit$}Yi R. Fung, \textsuperscript{$\diamondsuit$}Tuhin Charkaborty, \textsuperscript{$\spadesuit$}Hao Guo, \\
\textsuperscript{$\heartsuit$}\textbf{Owen Rambow}, \textsuperscript{$\diamondsuit$}\textbf{Smaranda Muresan}, \textsuperscript{$\spadesuit$}\textbf{Heng Ji}\\
  \textsuperscript{$\spadesuit$}University of Illinois Urbana-Champaign \\
  \textsuperscript{$\diamondsuit$}Columbia University, \textsuperscript{$\heartsuit$}Stony Brook University \\
  \texttt{\{yifung2,hengji\}@illinois.edu}} 

\begin{document}
\maketitle

\begin{abstract}
 Knowledge of norms is needed to understand and reason about acceptable behavior 
in human communication and interactions across sociocultural scenarios. Most computational research on norms has focused on a single culture, and manually built datasets, from non-conversational settings. We address these limitations by proposing a new framework, \textbf{\textsc{NormSage}}\footnote{Our code and data are available at the Github repo  \href{https://github.com/yrf1/NormSage}{here}.},  to automatically extract 
 culture-specific norms from multi-lingual conversations. \textbf{\textsc{NormSage}} uses GPT-3 prompting to 1) extract \emph{candidate norms} directly from conversations and 2) provide \emph{explainable self-verification} to ensure correctness and relevance. Comprehensive empirical results show the promise of our approach to extract high-quality culture-aware norms from multi-lingual conversations (English and Chinese), across several quality metrics. Further, our relevance verification can be extended to assess the adherence and violation of \textit{any} norm with respect to a conversation on-the-fly, along with textual explanation. \textbf{\textsc{NormSage}} achieves an AUC of 94.6\% in this grounding setup, with generated explanations matching human-written quality. 
\end{abstract}
\section{Introduction}
Norms are rules that embody the shared standards of behaviors amongst cultural groups and societies \cite{schwartz2012overview}. 
These include \textit{social conventions} ({\em e.g.,} \textcolor{gray}{it's good to shake hand with your opponent even if you lost}); \textit{behavior guidances} ({\em e.g.,} \textcolor{gray}{it's wrong to hurt a pet}); and \textit{general concepts} ({\em e.g.,} \textcolor{gray}{it's nice to be smart}) \cite{forbes2020social,ziems2022moral}. 
Knowledge of norms in general, and sociocultural norms in particular, is essential if we are to equip AI systems with capability to understand and reason about acceptable behavior in communication and interaction across cultures. 

Yet, research on computational modeling of social norms is still in its early stages. \textsc{Social-Chem-101} \cite{forbes2020social} and the \textsc{Moral Integrity Corpus} \cite{ziems2022moral} introduce large crowdsourced datasets of social norms and moral judgements conceptualized as free-text Rules-of-Thumb (RoT) grounded in given situations. While this has led to significant advances in computational modeling of social norms, there are two important limitations. First, while models trained on these  grounded RoTs can adapt to new situations, they often struggle to uncover new norms and to model richer context. 
Second, the norms covered in these studies are primarily Reddit-based and US-centric \cite{RedditUserStats}.    

\begin{figure*}[h!]
\includegraphics[trim={0 4cm 0 2cm},width=\textwidth]
{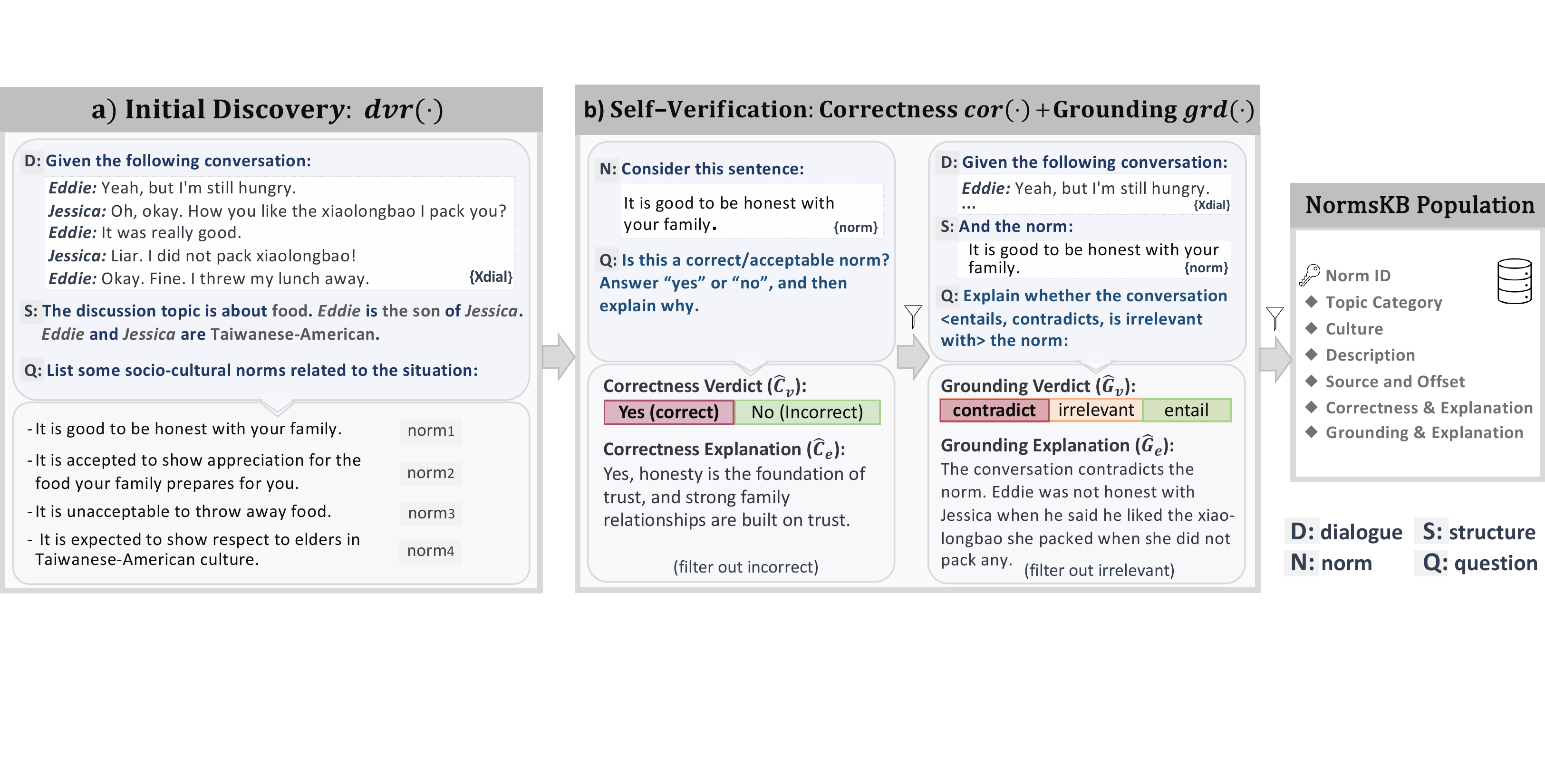}
\caption{The key idea of \textbf{\textsc{NormSage}} is prompting \& verification
for conversation-grounded norm discovery. Norm candidates are first discovered from dialogues, through the \textbf{\textit{dvr($\cdot$)}} operation. They are then checked for correctness and relevance, through \textbf{\textit{cor($\cdot$)}}  and \textbf{\textit{grd($\cdot$)}}, before populating into the knowledge base of norms, \textbf{NormsKB}.
}
\label{fig:1}
\vspace{-7pt}
\end{figure*}

In this paper, we propose to address current shortcomings with the following contributions:
\begin{itemize}[topsep=2pt,leftmargin=*]
\itemsep0em 
\item {\bf A new task: conversation-based multilingual and multi-cultural norm discovery (Sec \ref{sec:2}).} 
Norms manifest themselves in day-to-day conversations, either by being implicitly adhered to or violated. 
For example, we observe in Fig \ref{fig:1} input a conversation between a Taiwanese-American parent and son about \textit{"not wasting food"}. We define a new task aimed at extracting norms from conversation context, beyond single statements and question-answer pairs from recollected summaries. 
To support this task, we also introduce a collection of conversations in English and Chinese spanning multiple genres (e.g., TV shows, videos, text messages), topics, and cultures. 

\item {\bf \textsc{NormSage}, a zero-shot language model prompting and self-verification framework for norm discovery from conversations (Sec \ref{sec:3}).} 
First, we extract candidate norms using GPT-3 \cite{NEURIPS2020_1457c0d6} prompting that takes the conversational and social context into account (e.g., culture,  social relations, discussion topic, etc.), 
as shown in the Fig \ref{fig:1} 
\textbf{\textit{dvr($\cdot$)}} operation for initial norm \underline{d}isco\underline{v}e\underline{r}y. Second, to ensure norm correctness and relevance, we propose the idea of explanation-aware self-verification through prompting-based operations \textbf{\textit{cor($\cdot$)}} and \textbf{\textit{grd($\cdot$)}}. As shown in Fig \ref{fig:1}(b), our formulation here involves a question assessing the \underline{cor}rectness of a candidate norm, and another probing how the norm \underline{gr}oun\underline{d}s to its source dialogue. 
The entailment reasoning in \textbf{\textit{grd($\cdot$)}} can be further used to 
assess the adherence or violation of \textit{any} norm in a conversation, with textual explanation. 
\item {\bf Comprehensive evaluation setup (Sec \ref{sec:4})}. \quad 
We assess the quality of culture-aware norm extractions along several dimensions. Our method discovers significantly more relevant and insightful norms for conversations on-the-fly (selected as \textit{"best"} $\geq$70\% cases by human judgement) compared to baselines developed over Reddit-based rules-of-thumb annotations \cite{forbes2020social,ziems2022moral}. Additionally, the norms discovered from Chinese conversation exhibit limited performance difference compared to the norms discovered from English conversation in terms of relevance and correctness ($\Delta\leq5$\%). The culture-specific norms are also promising in quality, with humans being able to identify the culture from these norms at 80\% accuracy.
\item {\bf NormKB}, a knowledge base of over $20$k unique norms with correctness and grounding explanations, extracted from multi-lingual multi-cultural conversations.
\end{itemize}

\section{Task Formulation and Source of Data}\label{sec:2}\vspace{-0.15cm}
\subsection{Task Formulation}
\begin{table*}[ht!]
    \centering
    \setlength\extrarowheight{-1.5pt}
    \begin{tabular}{|c|c|c|c|c|c|}
        \hline
        & \small{\textbf{Source of Data}} & \small{\textbf{Lang}} & \small{\textbf{Topic}} & \small{\textbf{\# Tok}} & \small{\textbf{\# Trn}} \\
        \hline
        \parbox[t]{2mm}{\multirow{3}{*}{\rotatebox[origin=c]{90}{\footnotesize{\textbf{Single Culture~}}}}} & \small{Big Bang Theory (BBT)} & \small{EN} & \small{the life of nerdy, brilliant scientists in California} &  \small{29,682} & \small{3,468} \\ 
        & \small{Friends (F)} & \small{EN} & \small{Manhattan adult life, friendship, dating} & \small{26,197} & \small{2,849} \\ 
        & \small{How I Met Your Mother (HIMYM)} & \small{EN} & \small{romance centered around NYC} & \small{29,423} & \small{3,785}\\ 
        & \small{Grey's Anatomy (GA)} & \small{EN} & \small{medical environments centered in Seattle} &  \small{23,341} & \small{3,117} \\ 
        & \small{Castle (C)} & \small{EN} & \small{law and justice centered around NY police} & \small{38,880} & \small{4,142} \\ 
        \hline
        \parbox[t]{2mm}{\multirow{3}{*}{\rotatebox[origin=c]{90}{\footnotesize{\textbf{Cross-Culture~}}}}} 
        & \small{Fresh off the Boat (FOB)} & \small{EN} & \small{Taiwanese family in America} & \small{26,056} & \small{4,129} \\ 
        & \small{Never Have I Ever (NHIE)} & \small{EN} & \small{Indian girl coming-of-age in the US} & \small{39,847} & \small{5,637} \\ 
        & \small{Blackish (B)} & \small{EN} & \small{African American family in the suburbs} &  \small{33,993} & \small{5,103} \\ 
        & \small{Citizen Khan (CK)} & \small{EN} & \small{Pakistani family in Britain, patriarchism} &  \small{22,985} & \small{3,198} \\
        & \small{Outsourced (O)}  & \small{EN} & \small{American salesman, outsourced to India} &  \small{1,464} & \small{123} \\
        \hline & &&&&
        \\[-2.1ex] 
        \parbox[t]{2mm}{\multirow{3}{*}{\rotatebox[origin=c]{90}{\footnotesize{\textbf{MultiLing}}}}} 
        & \small{American Factory (AF)} & \small{EN,CN} & \small{Chinese company open factory in Ohio} & \small{10,840} & \small{1,138} \\
        & \small{Real-World Negotiations (RWN)} & \small{EN,CN} & \small{US-China talks, street vendor purchase talks} & \small{19,487} & \small{1,758} \\
        & \small{LDC2022E11 CCU TA1 Dev.} & \small{CN} & \small{text msg, phone calls, online videos} & \small{1.2M} & \small{102k} \\[1.4mm]
        \hline
        & \small{Total} & - & - &  \small{1.5M} & \small{140k} \\
        \hline 
    \end{tabular}
    \caption{We list out the sources of our raw data here, along with description of their language (EN - English, CN - Chinese), topic domain, and statistics (\# tok - number of spoken tokens, \# trn - number of dialogue turns).}
    \label{tab:dataset_details}
    \vspace{-10pt}
\end{table*}
\noindent We define the \textbf{conversation-based, multi-lingual multi-cultural norm discovery} problem as follows.  
Given a conversation scenario $(\textbf{X}\textsubscript{dial})$ in one of the pre-defined target languages ({\em e.g.}, English, Chinese), we aim to derive a set of relevant norms $\textbf{N}=\{\,\textbf{n}_{1}...\textbf{n}_{m} \}$ in English, for a 
 unified representation. If background information (\textbf{X}\textsubscript{b}) about the conversation context or speaker profile is available, such as from the Wikipedia \underline{s}ummary of a TV show, it can optionally be incorporated to enrich norm discovery. 
Because handling hallucination and achieving transparency are important for norm discovery, we introduce checking \textbf{norm correctness} and \textbf{grounding} as  supplementary \textbf{verification} tasks. We aim to filter out incorrect norms, by deriving a correctness verdict $\textbf{C}\textsubscript{v} \in \{\textbf{1}: yes, \textbf{-1}: no\}$, along with a confidence probability and natural language explanation. Additionally, we aim to filter out non-insightful norms, by deriving a grounding inference $\textbf{G}\textsubscript{v} \in \{\textbf{1}: entail, \textbf{0}: irrelevant, \textbf{-1}: contradict\}$, along with confidence probability and explanation. 

Our proposed task is innovative in several key aspects. It is the first task to define automatically discovering norms from dialogue data, which best reflects human communication on-the-fly. 
In addition, it is the first task on discovering multicultural norms from multilingual sources, which is then used to construct a \textbf{NormsKB} that can better represent diverse socioethnic or demographic groups. 
Finally, grounding the discovered norm with dialogue examples, confidence score, and natural language explanations enables our norm discovery approach to be explainable and self-supervised. 
This verification process can be used to check whether a conversation adheres or violates any given norm.  
\subsection{Source of Data}\vspace{-0.05cm}
In collecting source data for norm discovery from 
\newpage 
\noindent conversations, we seek data that involve dialogue exchanges mimicing or reflecting real-world communication. Secondly, the data should ideally span diverse topics and societal or cultural groups. In practice, it is generally difficult to obtain large-scale, in-the-wild data for norm discovery due to privacy concerns, as well as sparsity of interesting human interaction occurrences. TV shows and movies offer an interesting alternative, as they represent naturally occurring conversations often involving multicultural exchanges. We expand on the predominantly single-cultured TVQA dataset \cite{lei-etal-2018-tvqa}, and collect a set of TV and movies that cover different cultures, all of which are primarily in English. 
We also include several multi-lingual (Chinese and English) conversations from real-world chats, negotiations, and documentaries, to explore norm discovery adaptability in diverse data settings, as detailed in Table \ref{tab:dataset_details}. In particular, the LDC data is a large, compiled release of several thousand SMS/chat, phone conversations, and YouTube/Bilibili video scenarios \cite{LDC2022E18}. For each source of data, we collect the corresponding textual summary from Wikipedia, if available. Lastly, we preprocess dialogue transcripts into chunks (\textbf{X}\textsubscript{dial$_{1..N}$}) every $k=5$ lines as the individual data points in our task consideration. 

\section{\textbf{\textsc{NormSage}} Framework}\label{sec:3}

\subsection{\textbf{Initial Norm Discovery}}  
\begin{figure}[]
\includegraphics[scale=0.56]{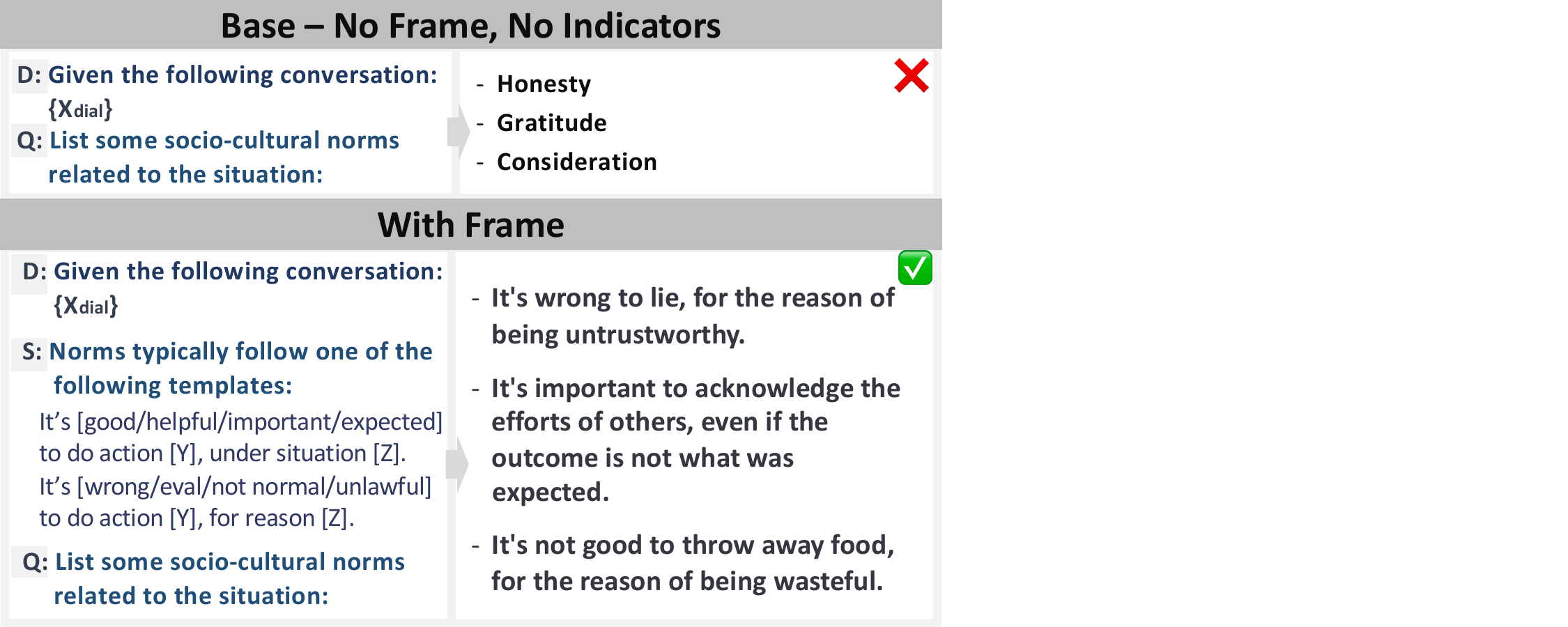} 
\caption{A comparison of \textbf{\textit{dvr($\cdot$)}}--\textsc{NormSage} in \textit{base} form and with \textit{framing} structure.} 
\label{fig:2}
\vspace{-9pt}
\end{figure}

\vspace{-0.05cm}
Pretrained language models store implicit knowledge about the world learned from large-scale text collected around the internet \cite{petroni-etal-2019-language,hierarchicalschema2023}. We frame conversation-based norm discovery as a series of natural language prompts, each with a directed question for the pretrained \textsc{GPT-3} Davinci language model to reason with its internal knowledge and generate an answer response.  
To \textbf{\underline{d}}isco\textbf{\underline{v}}e\textbf{\underline{r}} an \textit{initial} set of candidate norms from conversation data, we introduce the \textbf{\textit{dvr}}\textbf{($\cdot$)} operator, which concatenates \hl{\textbf{D}}, a template header describing the nature of the context data followed by the dialogue input \{$\textbf{X}_{\textbf{dial}_{i}}$\}, with \hl{\textbf{Q}}, a directed question describing the norm discovery task, as input for the PLM to generate response. We follow the general prompt development guideline of wording \textbf{D} and \textbf{Q} in clear and concise language. 
\paragraph{Structure Enhancement} A shortcoming observed in standard prompting is that the norms discovered may lack well-formedness and taxonomy for categorizing information specific to different cultures and topics. To encourage greater level of detail and structure in \textbf{\textit{dvr}}\textbf{($\cdot$)} outputs, we investigate adding to the prompt input: 

\begin{itemize}[nosep,leftmargin=3mm,label={}]
    \itemsep0em 
    \item \hl{\textbf{S}} -- a building block in the text template consisting of either frame\textbf{\underline{s}} defining the expected structure of norms (see Fig \ref{fig:2}), or cultural  indicator\textbf{\underline{s}} encouraging topic-specific taxonomy (see Fig \ref{fig:1}a). 
    These cultural indicators are extracted automatically through prompting on the background summary if the data is available. Further details of this process are included in \ref{A.1}.
\end{itemize}

\subsection{Self-Verification with Correctness Checking \& Explainable Grounding} 
For each discovered norm, we add a $\textbf{\textit{cor}}\textbf{($\cdot$)}$ operator to check the correctness of norms through prompting. This prompting operator follows the natural language template of: \textcolor{gray}{\textit{"Consider this sentence: \{\underline{$\textbf{n}_{i}$}\}. Is this a correct/acceptable social norm? Answer `\textbf{yes}' or `\textbf{no}', and then explain why."}}. The output from \textbf{\textit{cor}}\textbf{($\cdot$)} consists of both a correctness verdict, $\textbf{C}_{\textbf{v}}$ and a subsequent explanation, $\textbf{C}_{\textbf{e}}$, in a single natural language response generation (Fig \ref{fig:1}b). 
We further derive a confidence score ($\hat{\textbf{C}}_{\textbf{p}}$) for the correctness verdict, by normalizing the probability of token generation for $\textbf{C}_{\textbf{v}}=$`yes' with the probability of token generation for $\textbf{C}_{\textbf{v}}=$`no'. 
Norm candidates with correctness probability below a tunable threshold of $\theta=0.7$ are filtered out. 

As norms are subjective in nature and language models have the risk of hallucination in their output predictions, we further safeguard norm discovery with a \textbf{\textit{grd}}($\cdot$) operator for determining whether the hypothesized norm discovery can be groundable to its situation premise. We reformulate the explainable NLI setting \cite{NEURIPS2018_4c7a167b,Huang2022,factualerrorcorrection2023} for our new conversation-based norm grounding task, and devise the following natural language template: \textcolor{gray}{\textit{"Explain whether the conversation $<$\textbf{entails}, \textbf{contradicts}, or \textbf{is irrelevant with}$>$ the given norm"}}. The output from \textbf{\textit{grd}}($\cdot$) consists of the grounding verdict, $\textbf{G}_{\textbf{v}}$, along with the explanation, $\textbf{G}_{\textbf{e}}$. 
We get the grounding relevance score by normalizing the token probability being either \textit{"entail"} or \textit{"contradict"} in $\textbf{G}_{\textbf{v}}$ over the sum of probability of generating \textit{"entail"}, \textit{"contradict"}, or \textit{"irrelevant"}. Then, we filter out the norm candidates with grounding score below a  tunable threshold, $\gamma=0.6$.
\begin{table*}[ht]
    \setlength\extrarowheight{-2pt}
    \centering
    \begin{adjustbox}{width=\textwidth}
    \small
    \begin{tabular}{|l|ccccc|c|c|}
        \hline & &&&&&
        \\[-1.5ex]
         & \textbf{Relevance} & \textbf{Well-Formedness} &
         \textbf{Correctness} & \textbf{Insightfulness} &
         \textbf{Relatableness} & 
         \textbf{Best}
         \\
        \hline & &&&&&
        \\[-1.5ex]
        \textbf{\textsc{SocialChem}}\textsubscript{rtv} & 3.8 & 4.0 & 3.9 & 3.8 & 3.9 & 6.2\% \\
        \textbf{\textsc{NMT}}\textsubscript{gen} & 3.1 & 3.9 & 3.5 & 3.4 & 3.7 & 0.0\% \\ 
        \textbf{\textsc{MIC}}\textsubscript{rtv} & 3.3 & 3.1 & 3.6 & 3.4 & 3.2 & 7.1\% \\ 
        \textbf{\textsc{pMT}}\textsubscript{gen} & 2.2 & 3.2 & 3.1 & 3.0 & 2.5 & 0.0\% \\ 
        \textbf{\textsc{T0}}\textsubscript{pp} & 2.7 & 2.0 & 2.0 & 2.1 & 2.1 & 0.0\% \\ 
        \hdashline
        \textbf{\textsc{NormSage}} & 3.9 & 2.8 & 4.1 & 3.5 & 3.6 & 51.3\% \\        
        \textbf{\textsc{NormSage}}\textsuperscript{+} & \textbf{4.5}  & \textbf{4.5} & \textbf{4.6} & \textbf{4.2} & \textbf{4.7} & \textbf{70.4\%} \\
        - \textbf{\textsc{{\textsc{GPT-3}}}}
         & 3.0 & 2.8 & 2.8 & 2.8 & 3.6 & 25.5\% \\
         - \textbf{\textsc{NormSage}}\textsubscript{mini} & 3.8 & 4.2 & 3.7 & 3.4 & 4.0 & 11.9\% \\
        \hline 
    \end{tabular}
    \end{adjustbox}
    \caption{Likert scale (1-5) ratings across five quality metrics and the percentage of norm discovery outputs considered as \textbf{"best"} for a given dialogue scenario, averaged over 100 data samples.}
    \label{tab:norm_comparison}
    \vspace{-10pt}
\end{table*}

Finally, when populating \textbf{NormsKB} with new norm discoveries, we disregard norms that duplicate existing ones in the norms library. We flag norms as duplication if their \textsc{BERT}(n) embeddings \cite{devlin-etal-2019-bert} exceed a threshold of cosine similarity with any previously discovered norm. The threshold is empirically set to $\sigma=0.95$.

\section{Evaluation and Results}\label{sec:4}\vspace{-0.15cm}

We organize this section as follows. In Sec \ref{4.1}, we detail experiments on norm discovery from conversations, including multi-lingual and cross-culture scenarios. Then, in Sec \ref{4.2} and Sec \ref{4.3}, we analyze the effectiveness of self-verification mechanisms for norm discovery correctness, and relevance grounding between the norm and conversation, respectively. Finally, in Sec \ref{4.4}, we discuss the cost and resource contributions.

\subsection{Intrinsic Norm Discovery Evaluation}\label{4.1}\vspace{-0.05cm}
\paragraph{Baselines} While there has been no prior work on norm discovery based on conversations, we include the following baselines trained on a different data format, as valuable points of comparison. \textbf{\textsc{NMT}}\textsubscript{gen} is a \textsc{GPT2-XL} trained on \textsc{SocialChem101} \cite{forbes2020social}, while \textbf{\textsc{SocialChem}}\textsubscript{rtv} retrieves the most relevant \textsc{SocialChem101} rule-of-thumbs for a dialogue based on their embeddings encoded from pre-trained \textsc{BERT} \cite{devlin-etal-2019-bert}. We also consider \textbf{p\textsc{MT}}\textsubscript{gen}, which is a generator trained on the \textsc{Moral Integrity Corpus (MIC)} \cite{ziems2022moral}, and \textbf{\textsc{MIC}}\textsubscript{rtv}, which retrieves the most relevant \textsc{MIC} rules-of-thumb for a dialogue based on their embeddings encoded from pre-trained \textsc{BERT}. In addition, we include \textbf{\textsc{T0}\textsubscript{pp}}, a \textsc{T5} model 16x smaller than \textsc{GPT-3} that is trained on tasks formulated as natural language prompts \cite{sanh2022multitask}. 
In our proposed \textsc{GPT-3} prompting with self-verification approach, we consider the variant that utilizes structural enhancement in initial norm discovery as $\textbf{\textsc{NormSage}\textsuperscript{+}}$. We also include \textbf{\textsc{GPT-3}} prompting, without self-verification and without structure enhancement, as an additional baseline. Moreover, we
include an open and accessible alternative, $\textbf{\textsc{NormSage}}\textsubscript{mini}$, which consists of a \textsc{GPT2-XL} finetuned on the norm discovery outputs from \textbf{\textsc{NormSage}}. 
\\[5pt]
\textbf{Metrics} \quad We proposed to evaluate norm discovery from conversations along the dimensions of relevance, well-formedness, correctness, insightfulness, and relatableness, using a Likert scale of 1 ("awful") to 5 ("excellent"). Definitions and assessment guidelines of these metric categories are included in \ref{A:2.1}.
We also assess what percentage of norms discovered from the various methods are rated as \textit{\textbf{best}} overall, for each dialogue scenario. More than one norm can be considered as "best".
\\[5pt]
\textbf{Setting} \quad We crowdsource on Amazon Mechanical Turk for human assessment. Each HIT (``submit" task) consists of a dialogue scenario, an assessment metric, and sets of norms representing the norms discovered from each de-identified generation or retrieval approach.
Following crowdsourcing guidelines \cite{sheehan2018crowdsourcing}, we provide definitions and examples for each assessment metric. Workers undergo a qualification test ensuring a $\geq$95\% HIT rate, comprehension of task instructions, and native fluency in the conversation language. 
For assessing norms in cross-cultural (pairwise) scenarios, we ensure that the workers have basic familiarity with both cultures. Based on these worker selection criteria, we assign each example to ten distinct workers, and reject poor quality hits. 
Workers take 1-2 minutes per norm comparison task, and are rewarded a \$15 hourly rate. 
\\[5pt]
\noindent \textbf{Results} \quad Table \ref{tab:norm_comparison} shows our norm discovery intrinsic evaluation results on English conversation scenarios. 
As we can see, our proposed norm discovery approach, \textbf{\textsc{NormSage}}, outperforms baselines across all dimensions when enhanced with either frame or structured indicators. A major limitation of baseline approaches is poor portability to conversation domains. The performance of \textbf{\textsc{SocialChem}}\textsubscript{rtv} and \textbf{\textsc{MIC}}\textsubscript{rtv} shows that simply retrieving pre-annotated norms results in the norms being less relevant and insightful for new conversations. Compared to the retrieval baselines, the generation baselines, \textbf{\textsc{NMT}}\textsubscript{gen} and \textbf{\textsc{pMT}}\textsubscript{gen}, perform even worse. This suggests that the domain gap in situation context between curated Reddit post headers (previous approaches) and in-situ conversations (current task) poses an even greater bottleneck for norm discovery here. 
\textbf{\textsc{NormSage}} overcomes challenges in domain portability through operationalizing zero-shot language model prompting for conversation reasoning and norm discovery. 

 \subsubsection{Multi-Lingual Norm Discovery}
\begin{figure}[h!]
\includegraphics[scale=0.248]{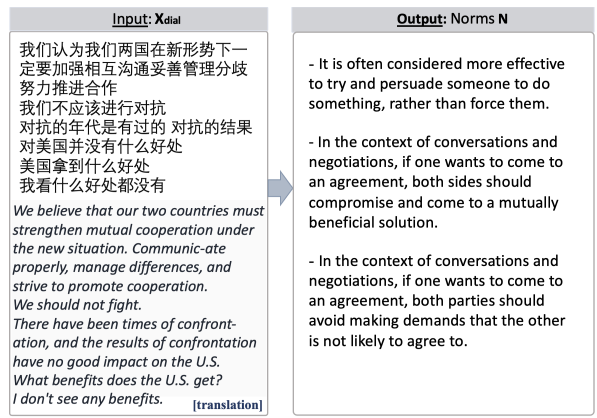}
\caption{Example illustration of \textbf{\textsc{NormSAGE}} norm discovery results on Chinese conversations, demonstrating our method's multi-lingual capability beyond English data.}
\label{fig:3}
\end{figure}

As a case study of the multi-lingual task setting, we investigate norm discovery performance on Chinese conversations, since Chinese is a widely spoken and studied language separate from the western language systems. As visualized in Fig \ref{fig:3}, the norms discovered from Chinese conversations are high-quality in detail and correctness. 
\begin{table*}[ht!]
    \setlength\extrarowheight{-1.7pt}
    \makebox[\textwidth]{
    \begin{tabular}{|p{13.65cm}|p{1.55cm}|}
        \hline  
        \textbf{\small{Culture-Specific Norm Discoveries}} & ~~~\small{\textbf{Source}} \\
        \hline  %
        \small{In \textbf{Pakistani} culture, it is not uncommon for the bride and groom to not meet each other until wedding day.$^{\bullet}$} 
        & ~~~~~~\small{CK} \\
        \small{In \textbf{India}, it is considered polite to always offer food and drink to guests, even if they decline.$^{\diamond}$ 
        } & ~~~~~~~\small{O} \\
        \small{In \textbf{Taiwanese} culture, it is more common to have a heavier lunch, such as rice and vegetables.$^{\circ}$ 
        } & ~~~~~\small{FOB} \\
        \small{In \textbf{British} culture, it is normal for the bride and groom to meet each other before wedding.$^{\bullet}$} & \small{~~~~~~~~CK} \\ 
        \small{In \textbf{American} culture, it is common to have a light lunch, such as a salad or sandwich.$^{\circ}$} & ~~~~~~~\small{O} \\
        \small{In \textbf{America}, it is more common to just let guests decline if they don't want anything.$^{\diamond}$} & ~~~~~\small{FOB} \\ 
        \hline
    \end{tabular}
    }
    
    \caption{Visualization of some examples in culture-specific norm discovery. We denote the pairs of contrasting norms across cultures with special symbols ($\diamond,\circ,\bullet$). 
    }
    \label{tab:cross_culture_comparison_visualization_examples}
    \vspace{-8pt}
\end{table*}

We also perform an investigation on the quality of norms discovered from Chinese conversations compared to norms discovered from English conversations. We measure the stand-alone correctness and insightfulness of norms discovered from multi-lingual setting, on a 1-5 Likert scale. 
The results in Table \ref{tab:multi_ling_norm_discovery_analysis} indicate that NormSage exhibits relative consistency in norm discovery across both English and Chinese conversations, with $\Delta\leq$9.5\% observed in terms of relevance and correctness. Interestingly, norms discovered from English data receive slightly higher rating for insightfulness ($\leq$9.5\%) but lower ratings for relatableness ($\leq$2.1\%), potentially due to the dialogue nature. The English sources involve movies and shows, which tend to be more creative (insightful) and less formal (biased dialogues may lower norm correctness). Finally, our prompting-based method for norm discovery from conversations can be applied to other languages that the large language model backbone, \textsc{GPT-3}, has been pretrained on.
 \begin{table}[h!]
    \centering
    \begin{adjustbox}{width=\columnwidth}
    \begin{tabular}{|c|c|c|c||}
        \hline 
        \small{\textbf{}} & \small{\textbf{CN Conv. Norm}} & \small{\textbf{$\Delta\%$ w. EN Conv. Norm}} \\
        \hline 
        \small{\textbf{Relevance}} &  4.3 & -4.4 \\
        \small{\textbf{Well-Formedness}} & 4.8 & +6.6 \\
        \small{\textbf{Correctness}} & 4.6 & 0.0 \\
        \small{\textbf{Insightfulness}} & 3.8 & -9.5 \\
        \small{\textbf{Relatableness}} & 4.8 & +2.1 \\
        \hline 
    \end{tabular}
    \end{adjustbox}
    \caption{Likert-scale (1--5) rating of norms discovered from Chinese (CN) conversation, and comparison with norms discovered from English (EN) data.}
    \label{tab:multi_ling_norm_discovery_analysis}
\end{table}

 \subsubsection{Culture-Specific Norm Discovery} 
\begin{table}[b!]
    \centering
    \small
    \begin{tabular}{|c|c|c|c|c|}
        \hline &&&
        \\[-2.3ex]
        \textbf{Culture Comparison} & \textbf{Source} & \textbf{Id} & \textbf{Rating} \\
        \hline  &&&
        \\[-2.3ex]
        \textbf{American} vs \textbf{Taiwanese} & FOB & 74 & 4.2 \\[0.5mm]
        \textbf{American} vs \textbf{Indian} & O & 82 & 4.0 \\[0.5mm]
        \textbf{Western} vs \textbf{Muslim} & CK & 91 & 4.5 \\[0.5mm]
        \makecell{\textbf{African American vs} \\ \textbf{Caucasian American}} &  B & 73 & 4.0 \\[0.5mm]
        \hline 
        \textbf{Average} & - & 80 & 4.1 \\
        \hline 
    \end{tabular}
    \caption{We evaluate culture-specific norm discovery via culture \textbf{\underline{id}}entification accuracy (\%) between pairs of norms and also Likert scale \textbf{\underline{rating}} (1--5) on the norms.} 
\label{tab:cross_culture_norm_discovery_analysis}
\vspace{-8pt}
\end{table}
 \noindent To evaluate the correctness of culture-specific norms, we design a pairwise culture comparison setting. Specifically, we run a pretrained \textbf{\textsc{bart-large-mnli}} model \cite{lewis-etal-2020-bart,N18-1101} on pairs of norm from different cultures, and randomly select 10 pairs of norm that are determined as "contradiction" with each other, for each of the cross-culture scenarios in our dataset ({\em e.g,} \textit{Fresh off the Boat}, \textit{Outsourced}, etc.). Then, we mask the culture identities in the pairs of norm, and ask human annotators familiar with both cultures to identity which culture each norm belongs to from binary options. The results, as shown in Table~\ref{tab:cross_culture_norm_discovery_analysis}, indicate that the culture-specific norms discovered from \textbf{\textsc{NormSage}\textsuperscript{+}} are promising, with human annotators achieving 80\% identification accuracy. The inter-annotator agreement for the culture-identification task, as measured by Fleiss Kappa, is 60.8\%, which signals moderately good agreement. Some of the error cases in culture comparison of norm discoveries may be due to subjectivity of the assessment task (for example, norms about whether the Caucasian American or African American cultural group is more likely to discuss controversial topic).

\subsection{Extrinsic Evaluation on Norm Correctness Self-Verification}\label{4.2}
\paragraph{Baselines} We compare \textbf{\textit{cor}($\cdot$)-\small{\textsc{NormSage}}}, the mechanism which checks for correctness in our proposed norm discovery approach, with:
\begin{itemize}[nosep,leftmargin=*]
    \itemsep0em 
    \item \textbf{\textsc{Bart--SocialChem--ag}}: We finetune a \textsc{bart-large} model \cite{lewis-etal-2020-bart} to perform regression and predict \textit{anticipated agreement} scores in the \textsc{Social-Chem-101} dataset \cite{forbes2020social}, which reports on a 1--5 scale how much people agree with the rules-of-thumb. 
    \item \textbf{\textsc{T0}\textsubscript{pp}}: For prompt template, we follow the same principles as for \textbf{\textit{cor}($\cdot$)-\small{\textsc{NormSage}}} to first introduce the norm and then ask a directed question about its correctness. 
\end{itemize}
For reproducibility in this classification task, we set temperature to 0 during prompting generation.
\\[5pt]
\textbf{Metrics} \quad Norm correctness verification is a two-class classification problem. We measure performance in terms of classification accuracy (\textbf{Acc}) and area under the ROC-curve (\textbf{AUC}). We also assess the textual explanations of correctness predictions through human rating on a 1--5 Likert scale. 
\\[5pt]
\textbf{Data Setting} \quad We randomly select a set of norms discovered from \textbf{\textit{dvr}($\cdot$)-\small{\textsc{NormSage}}}, and ask crowd workers (paid \$15/hr) to provide correctness labels and explanations. We keep the data in which the correctness label has been agreed upon by at least three annotators. In total, we derive a set of 100 norms that are class-balanced with half "correct" and half "incorrect" labels.
\\[5pt]
\textbf{Results} \quad Table \ref{tab:norm_correctness_comparison} shows the results of norm correctness verification. The zero-shot prompting approaches, \textbf{\textit{cor}($\cdot$)-\small{\textsc{NormSage}}} and \textbf{\textsc{T0}}\textsubscript{pp}, achieve the best performance for this task. Interestingly, we observe that \textbf{\textsc{T0}}\textsubscript{pp} has a higher Acc and AUC score than \textbf{\textit{cor}($\cdot$)-\small{\textsc{NormSage}}}. We believe this may be due to the fact that the norms have been generated by the same underlying \textsc{GPT-3} language model backbone now used to verify the data. As a result, the language model may tend to associate the semantic content or linguistic artifacts in the norm as correct. However, \textbf{\textit{cor}($\cdot$)-\small{\textsc{NormSage}}} remains a competitive method for verifying norm correctness and has the added advantage of generating high-quality explanations for its predictions, enabling transparent reasoning.  
In contrast, \textbf{\textsc{T0}}\textsubscript{pp} prompting cannot generate reasonable explanation on norm correctness beyond circular reasoning. Finally, even though \textbf{\textsc{Bart-SocialChem-ag}} is finetuned on \textsc{Social-Chem-101} \textit{"agreement"} annotations, it does not perform as well in our norm correctness task. This is likely due to the lack of culture-specific rules-of-thumb annotated in the \textsc{Social-Chem-101} dataset to train a robust correctness verification model. 
\begin{table}[hbt!]
    \centering
    \setlength\extrarowheight{-1.5pt}
    \begin{adjustbox}{width=\columnwidth}
    \small
    \begin{tabular}{|c|c|c|c|}
        \hline &&&
        \\[-1.7ex]
        & \textbf{Acc} & \textbf{AUC} & \textbf{Expl} \\
        \hline  &&&
        \\[-1.7ex]
        \textbf{\textsc{Bart}}-\textbf{\textsc{SocialChem--ag}} & 75 & 78.0 & N/A \\
        \textbf{\textsc{T0}\textsubscript{pp}} & 82 & 92.0 & 1.0 \\
        \textbf{\textit{cor}}($\cdot$)-\textbf{\textsc{NormSage}} & 81 & 85.3 & 4.2 \\
        \hline 
    \end{tabular}
    \end{adjustbox}
    \caption{Classification results (\%) and Likert rating (1--5) on the generated explanations for norm correctness.}
    \label{tab:norm_correctness_comparison}
    \vspace{-12pt}
\end{table}

\begin{table*}[hbt!]
\begin{adjustbox}{width=\textwidth}
    \begin{tabular}{|p{8.8cm}|p{2.2cm}|p{2.8cm}|p{0.98cm}|}
        \hline 
        \centering{\small{\textbf{Dialogue Situation}}} & \centerline{\small{\textbf{Discovered Norms}}} & \centerline{\small{\textbf{Grounding Explanation}}} & \small{\textbf{Label}} \\[-0.5cm]
        \hline 
        \small{\textbf{Dave}: No. Definitely booked.} & \multirow{6}*{\parbox{2.2cm}{\small{It's important to listen to others and give them a chance to speak.}}} & \multirow{6}*{\parbox{2.2cm}{\small{Mr. Khan is \colorbox{lightgray}{not listening} to Dave and he is \colorbox{lightgray}{not giving Dave a} \colorbox{lightgray}{chance to speak}.}}} & \multirow{6}*{\quad -1} \\
        \small{\underline{\textbf{Mr. Khan}}: \colorbox{lightgray}{What?!} Do know who I am? Hello! Mr Khan, community leader! Next President of Sparkhill Pakistani Business Association!} & & &  \\
        \small{\textbf{Dave}: I'm sorry} & & & \\
        \small{\underline{\textbf{Mr. Khan}}: Right, \colorbox{lightgray}{that's it}. I want to speak to the proper manager.} & & & \\
        \small{\textbf{Dave}: I am the property manager.} & & & \\
        \hline
        \small{\textbf{Beckett:} Sure I can, until a jury tells me otherwise.} & \multirow{4.5}*{\parbox{2.2cm}{\small{It is generally considered impolite to make lewd comments.}}} & \multirow{4}*{\parbox{2.5cm}{\small{What's spoken by Creason is irrelevant with the norm.}}} & \multirow{4}*{\quad 0}\\
        \small{\textbf{Creason}: You are wasting my time. Detective, look, I told you exactly what I was doing last night.} & & & \\
        \small{\textbf{Beckett}: Right. You were at the club. They said that you made quite the entrance [...]} & & & \\
        \hline
        \small{\textbf{Jessica}: [...] Well, those kids, they just don't know, that's all. It just -- it just take time to get used to something different.} 
        & \multirow{7}*{\parbox{2.2cm}{\small{It is also considered polite to try to make the best of a situation, even if you do not like it}}} & \multirow{7}*{\parbox{2.9cm}{\small{The mother is trying to \colorbox{lightgray}{make the best of the} \colorbox{lightgray}{situation} even though \colorbox{lightgray}{she does not like it}}}}& \multirow{7}*{\quad 1}\\ 
        \small{\textbf{Eddie}: I hate it here! I want to go back to D.C.} &  &  &\\
        \small{\underline{\textbf{Jessica}}: Eddie, that's not possible. We are here now. We have to \colorbox{lightgray}{make the best of it}. Like I am doing with this neighbor woman. You think I like pretending Samantha isn't carrying a baggie of dog poops in her hand? No! \colorbox{lightgray}{I don't like this}! ...But I am trying!} &  &  &  \\ 
        \hline
    \end{tabular}
    \end{adjustbox}
    
    \caption{{\color{red}
    }Norm grounding example results, randomly sampled for each class from \textit{\{-1: Contradict, 0: Irrelevant, 1: Entail\}}. We \underline{underline} the utterance-level provenance of the input, where entailment or contradiction is found. 
    }
    \label{tab:grounding_visualization}
    \vspace{-8pt}
\end{table*}
\subsection{Extrinsic Evaluation on Norm Grounding}\label{4.3}\vspace{-0.05cm}
Norm grounding is utilized in the self-verification process of our norm discovery, as  \textbf{\textit{grd}}($\cdot$)--\textbf{\textsc{NormSage}}. This subtask extends to identifying norm adherence and violation in conversations, and stands as an important downstream application. 
\\[5pt]
\textbf{Baselines} \quad We compare \textbf{\textit{grd}}($\cdot$)-\textbf{\textsc{NormSage}} with existing NLI-based approaches, including \textbf{\textsc{BART-MNLI}}, a \textsc{BART-large} model \cite{lewis-etal-2020-bart} pretrained on Multi-genre NLI \cite{N18-1101}; \textbf{\textsc{BART-DialNLI}}, a \textsc{BART-large} pretrained on Dialogue NLI \cite{welleck2018dialogue}; and \textbf{\textsc{T5}}-e\textbf{\textsc{SNLI}}, a \textsc{T5} \cite{raffel2020exploring} pretrained on explainable NLI \cite{NEURIPS2018_4c7a167b}. We further consider  $\textsc{\textbf{T0}}\textsubscript{pp}$, introduced in Sec \ref{4.1}. For its prompt template here, we follow the same principles as for \textbf{\textit{grd($\cdot$)}-\textsc{NormSage}} to introduce the dialogue context and norm, and question about its adherence/violation relevance. We also evaluate training a \textsc{bart-large} model on the grounding outputs of \textbf{\textsc{NormSage}}, as \textbf{\textit{grd($\cdot$)}-\textsc{NormSage}\textsubscript{mini}}. Similar to Sec \ref{4.2}, we remove all model randomness. 
\\[5pt]
\textbf{Metrics} \quad Similar to the norm correctness task, we evaluate norm grounding in terms of classification accuracy (\textbf{Acc}) and area under the ROC-curve (\textbf{AUC}). For both metrics, we consider a two-class scenario to determine whether a norm is \textit{relevant} (either entailed or contradicted) or \textit{irrelevant} to a dialogue. We also consider a finer-grained three-class scenario on whether the norm is \textit{entailed}, \textit{contradicted}, or \textit{irrelevant}.
As one speaker may violate a norm while another points out adherence issues, we evaluate three-class norm grounding conditioned on speaker localization ({\em i.e.,} for the dialogue of a given speaker), if speaker identity is available. 
Finally, we use human assessment to evaluate grounding prediction explanations on a 1--5 Likert scale.
\\[5pt]
\textbf{Data Setting} \quad 
We randomly sampled dialogue chunks and norm candidates to create a set of 100 data annotations with class-balanced <\textit{contradict}, \textit{irrelevant}, or \textit{entail}> labels and corresponding explanations. The annotations include speaker identities for dialogues with known speakers. During three-class inference, we consider two data input variants for each method, with and without speaker localization, to investigate whether deeper differentiating of who violates and who adheres to a norm helps. Speaker localization is incorporated into NLI classification models by using only the dialogue lines from the speaker of interest as input. It is incorporated into the prompting-based \textbf{\textit{grd($\cdot$)}-\textsc{NormSage}} and  \textbf{\textsc{T0}}\textsubscript{pp} approaches via a question in the prompt template, "Explain whether what's spoken by [speaker ID] $<$\textit{entails}, \textit{contradicts}, \textit{is irrelevant with}$>$ the norm". We report three-class grounding results utilizing the data setting, with or without speaker localization, that maximizes each model's performance and clarity the details next. 
\\[5pt]
\noindent \textbf{Results} \quad Speaker localization does not benefit NLI classification models for two-class norm grounding, possibly due to a domain shift in semantics after isolating a speaker's dialogue, but it plays a crucial role in prompting-based approaches for three-class norm grounding. In particular, \textbf{\textit{grd($\cdot$)}-\textsc{NormSage}} significantly outperforms all other baselines in norm grounding, as shown in Table \ref{tab:norm_grounding_comparison}. For three-class norm grounding, it achieves accuracies of 68\% without and 80\% with speaker localization. The findings suggest that speaker localization can allow for a fairer evaluation accounting for differences in norm grounding across speakers, and help models reason over the perspective and context of a speaker. 
We further highlight that \textbf{\textit{grd($\cdot$)}-\textsc{NormSage}} generates easy-to-follow and insightful norm grounding explanations that are preferred over human-written ones in around 40\% of cases. Examples of the grounding and explanation of norms discovered from conversations are visualized in Table \ref{tab:grounding_visualization}. It is worth noting that even though the dialogue inputs may contain improper grammar, involving noisy or informally styled textual dialogue, this is exactly the nature of real-world conversations which makes our novel norm discovery on-the-fly task setting introduced unique and challenging. Despite the noisy nature of dialogue inputs, Table \ref{tab:grounding_visualization} shows that our large language model prompting mechanism can still effectively reason over the <entailment, irrelevance, contradiction> nature of grounding between a given dialogue input and a given norm with proper explanations.
\begin{table}[h!]
    \centering
    \begin{adjustbox}{width=\columnwidth}
    \small
    \begin{tabular}{|c|c|c|c|c|c|}
        \hline 
        & \multicolumn{2}{c|}{\textbf{(2-Class)}} & \multicolumn{2}{c|}{\textbf{(3-Class)}} &  \\
        \cline{2-5}
        & \textbf{ACC} & \textbf{AUC} & \textbf{ACC} & \textbf{AUC} & \textbf{Expl} \\
        \hline   &&&&&
        \\[-2.1ex]
        \textbf{\textsc{Bart}}-\textbf{\textsc{MNLI}} & 36 & 31.5 & 32 & 27.3 & N/A \\
        \textbf{\textsc{T5}}-e\textbf{\textsc{SNLI}} & 50 & 22.9 & 33 & 20.0 & 1.92 \\
        \textbf{\textsc{BART}}-\textbf{\textsc{DialNLI}} & 67 & 70.1 & 33 & 44.9 & N/A \\
        \textbf{\textsc{T0}\textsubscript{pp}} & 67 & 94.2 & 29 & 34.1 & 1.3 \\
        \textbf{\textit{grd($\cdot$)}-\textsc{NormSage}\textsubscript{mini}} & 69 & 75.3 & 59 & 53.8 & 3.5 \\
        \textbf{\textit{grd($\cdot$)}-\textsc{NormSage}} & 79 & 91.6 & \textbf{80} & \textbf{92.7} & 4.1 \\
        \hline 
    \end{tabular}
    \end{adjustbox}
    \caption{Classification results (\%) on norm grounding, along with rating (1--5) on the generated explanations.}
    \label{tab:norm_grounding_comparison}
    \vspace{-8pt}
\end{table}

\subsection{Resource Contribution}\label{4.4}\vspace{-0.05cm}
We discovered over 20,500 unique norms, of which 1,250 are culture-specific. On average, $\textsc{\textbf{NormSage}}$ discovers norms at a rate of 8.6 seconds per dialogue, and performs norm grounding at a rate of 3.8 seconds per dialogue. This process is done through OpenAI API, which may fluctuate slightly due to internet speed, but has the benefit of requiring no GPU usage from the researcher's end. 
The $\textsc{\textbf{NormSage}}$ norm discovery process is over 10x faster the human annotation efforts. The cost of OpenAI GPT-3 API access is currently \$0.06 per 1k tokens, which is still less expensive than human annotation efforts. While quality may be a factor of difference, we showed that \textsc{\textbf{NormSage}}\textsubscript{mini}, a variant of the proposed approach trained on silver-standard \textsc{NormSage} outputs, perform competitively for norm discovery and verification tasks.
\section{Related Work}\vspace{-0.05cm}
The domain of norms is closely related to behavioral psychology and moral judgement. Early studies investigated the pragmatic cooperative principles \cite{grice1975logic}, politeness implicatures \cite{kallia2004linguistic}, and relationship between norms and law \cite{eric2009normlaw} governing human behavior. As judgements of behavior are communicated through linguistics, \cite{Graham2009MoralFound} introduced a lexicon of evocative words based on moral foundation theory, which later attempts utilize for predicting the moral value from text messages~\cite{Lin2018c,Mooijman2018}. Recent approaches explore modeling moral and ethical judgement of real-life anecdotes from Reddit \cite{emelin2020moral,sap2019social,lourie2021scruples,botzer2022analysis}, with \textsc{Delphi} \cite{jiang2021delphi} unifying the moral judgement prediction on these related benchmarks. Related is another line of work modeling legal judgement on judicial corpora \cite{chalkidis-etal-2022-lexglue}. %

Norm discovery is a unique, emerging task, which aims to catalogue the underlying principles behind behavioral judgements, and can be seen as similar to distilling reactions, explanations, and implications from situations \cite{vu-etal-2014-acquiring,10.5555/3016100.3016313,rashkin-etal-2018-event2mind,sap2019atomic}. Separate from explicit information extraction \cite{wen-etal-2021-resin}, such as the extraction of events/entities/relations which may tend to be overly granular and situation-specific, norm discovery involves inferring the underlying stand-alone abstracted social or cultural rules. Towards this end, \citet{forbes2020social,ziems2022moral} are the main existing norm discovery approaches. Each presents a large-scale catalogue of manually curated rule-of-thumbs from Reddit post headers, and trains a language model to generate rule-of-thumbs based on this data. In contrast, our work focuses on norm discovery from conversations on-the-fly and without needing manual curation.

Modeling the social and moral dynamics in human interaction and communication have diverse applications, such as the detection of cyberbullying \citep{van-hee-etal-2015-detection} and hate speech \citep{mathew2021hatexplain}, detoxification~\cite{LMSwitch2023}, debiasing~\cite{debiasinggradient2023,debiasing2023b,Adept2023}, model alignment \cite{ouyang2022training,dong2023raft}, bipartisan news framing \cite{fulgoni-etal-2016-empirical}, social media understanding \cite{sun2023}, 
emotions analysis \cite{zadeh2018multimodal,yu-etal-2020-ch}, and situational report generation \cite{reddy2023smartbook}, amongst other interdisciplinary human-centered NLP applications \cite{Li2023NewPlayground}. In particular, discovering norms is essential for \textit{explicitly} detecting norm adherence and violations instances (our work), as well as \textit{implicitly} guiding dialogues \cite{ziems2022moral}. From a technical perspective, our norm discovery approach based on language model prompting and knowledge elicitation can be seen as a form of prompt engineering \cite{le-scao-rush-2021-many}, where we prefix a question with an elaborated scene, with the underlying core intuition based on leveraging the zero-shot question answering capability of language models \cite{gangi-reddy-etal-2022-zero}. 
The norm grounding with explanation task is intuitively similar to the explainable natural language inference problem setting \cite{welleck2018dialogue,wiegreffe2021reframing}. Our proposed framework, \textbf{\textsc{NormSage}}, achieves norm discovery and grounding without intensive prompt-tuning \cite{jiang-etal-2021-know} or finetuning \cite{forbes2020social,ziems2022moral}.

\section{Conclusions}
We introduced a new NLP paradigm of guiding cross-culture communication with on-the-fly sociocultural-aware norm discovery, violation detection, and explanations. This conversation-based norm discovery and grounding problem goes beyond the US-centric data predominent in prior work. 
To address the new challenges, we present a framework called \textbf{\textsc{NormSage}} that leverages knowledge elicitation from large language model (LM) prompting and incorporates novel self-verification mechanisms. Our approach surpasses baselines by improving the discovery of dialogue norms across diverse social and cultural groups, including multi-lingual conversations and culture-specific norms, while providing natural language explanations for interpretable norm discovery and violation detection. Our work has broad impacts as it empowers the discovery of norms which may differ from each other across cultures, thus enabling individuals to navigate communication barriers more effectively and find common grounds that avoid norm violations. 
\section*{Limitations} 
Our norm discovery process makes use of the \textsc{GPT-3} from OpenAI\footnote{\url{https://openai.com/api/}} as a strong pre-trained language model to elicit groundable knowledge about the rules and judgements of acceptable behavior from human dialogue interactions. We recognize that norms may shift with context over time. Our discovery of norms applies to the time period that aligns with the conversation scenario in which a norm is discovered from. While the newer GPT-4 has now been released, we chose to stick with GPT-3 in our experiments due to the tight rate limit of GPT-4, currently capped at 25 request calls every 3 hours. 
We further point out that the \textsc{GPT-3} model acquired its implicit knowledge from ultra large-scale data, and has added in mechanisms to address bias \cite{solaiman2021process}. Nevertheless, all computational models still come with a risk of potential bias. For example, the behavior of some closed source models, such as ChatGPT, is not always guaranteed to be consistent over time \cite{chen2023chatgpts}. Moreover, explanations generated by language models may not always entail the models’ predictions nor be factually grounded in the input \cite{NEURIPS2022_c4025018}. We encourage researchers and practitioners to exercise caution and check-guards in their endeavors. 

\section*{Ethical Considerations} 
We recognize that the automatic generation of norms and judgements could be seen as normative and authoritative \cite{talat2021word,ziems2022moral}. Here, we emphasize that the discovered norms are not intended to form a global and universally binding ethical system, but rather to provide a set of discrete intuitions and principles to help differentially explain the underlying assumptions that exist latently. The present work supports an explainable system to automatically discover norms from conversations on-the-fly, and verify whether the discovered norms can be sufficiently grounded to the data source, as well as the (entail vs. contradict) relation characteristic. Currently, we do not specifically distinguish between different levels in the cultural hierarchy (such as the relationship between Cantonese and Chinese culture), and resort to cultural references in the source data. It is also important to note that some automatically discovered norms might be viewed as stereotypical. Our transparent and flexible system should be seen as a human-AI collaboration tool that domain experts interface with, facilitating the moderation efforts.

\paragraph{Privacy} We abide by privacy guidelines for data used in our research study. For example, the SMS conversation data in the LDC source of Tab \ref{tab:dataset_details} has been collected through previous voluntary paid participation program such as BOLT \cite{song2014collecting}. Furthermore, the conversation data involved are utilized for research purpose only. 
\paragraph{Risks and Mitigations} Our task involves source data that may contain explicit conversations about race, gender, religion, etc. We recognize the emotional burden that this presents to annotators \cite{roberts2016commercial}. In mitigation, we include the following content warning in the header of each task: \textit{This HIT may contain text that disturbs some workers. If at any point you do not feel comfortable, please feel free to skip the HIT or take a break}. The study has been thoroughly reviewed and approved by a national level internal review board. The resources and findings presented in this work are intended for research purposes only. To ensure proper, rather than malicious, application of dual-use technology, we require users of our norm discovery data to complete a Data Usage Agreement that we link in our project repository. We also intend to make our software available as open source for public auditing, and explore measures to protect vulnerable groups.

\section*{Acknowledgement}
This research is based upon work supported by U.S. DARPA CCU Program No. HR001122C0034. The opinions, views and conclusions contained herein are those of the authors and should not be interpreted as necessarily representing the official policies, either expressed or implied, of DARPA or the U.S. Government. The U.S. Government is authorized to reproduce and distribute reprints for governmental purposes notwithstanding any copyright annotation therein.

\bibliography{anthology,custom}

\begin{thebibliography}{60}
\expandafter\ifx\csname natexlab\endcsname\relax\def\natexlab#1{#1}\fi

\bibitem[{LDC(2022)}]{LDC2022E18}
 2022.
\newblock {CCU} {TA}1 {M}andarin/{C}hinese {D}evelopment {A}nnotation {LDC}2022{E}18.
\newblock Web Download.

\bibitem[{Bianchi(2022)}]{RedditUserStats}
Tiago Bianchi. 2022.
\newblock \href {https://www.statista.com/statistics/325144/reddit-global-active-user-distribution/} {Regional distribution of desktop traffic to reddit.com as of may 2022 by country}.
\newblock Accessed: 2023-01-31.

\bibitem[{Botzer et~al.(2022)Botzer, Gu, and Weninger}]{botzer2022analysis}
Nicholas Botzer, Shawn Gu, and Tim Weninger. 2022.
\newblock Analysis of moral judgment on reddit.
\newblock \emph{IEEE Transactions on Computational Social Systems}.

\bibitem[{Brown et~al.(2020)Brown, Mann, Ryder, Subbiah, Kaplan, Dhariwal, Neelakantan, Shyam, Sastry, Askell, Agarwal, Herbert-Voss, Krueger, Henighan, Child, Ramesh, Ziegler, Wu, Winter, Hesse, Chen, Sigler, Litwin, Gray, Chess, Clark, Berner, McCandlish, Radford, Sutskever, and Amodei}]{NEURIPS2020_1457c0d6}
Tom Brown, Benjamin Mann, Nick Ryder, Melanie Subbiah, Jared~D Kaplan, Prafulla Dhariwal, Arvind Neelakantan, Pranav Shyam, Girish Sastry, Amanda Askell, Sandhini Agarwal, Ariel Herbert-Voss, Gretchen Krueger, Tom Henighan, Rewon Child, Aditya Ramesh, Daniel Ziegler, Jeffrey Wu, Clemens Winter, Chris Hesse, Mark Chen, Eric Sigler, Mateusz Litwin, Scott Gray, Benjamin Chess, Jack Clark, Christopher Berner, Sam McCandlish, Alec Radford, Ilya Sutskever, and Dario Amodei. 2020.
\newblock \href {https://proceedings.neurips.cc/paper/2020/file/1457c0d6bfcb4967418bfb8ac142f64a-Paper.pdf} {Language models are few-shot learners}.
\newblock In \emph{Advances in Neural Information Processing Systems}, volume~33, pages 1877--1901. Curran Associates, Inc.

\bibitem[{Camburu et~al.(2018)Camburu, Rockt\"{a}schel, Lukasiewicz, and Blunsom}]{NEURIPS2018_4c7a167b}
Oana-Maria Camburu, Tim Rockt\"{a}schel, Thomas Lukasiewicz, and Phil Blunsom. 2018.
\newblock \href {https://proceedings.neurips.cc/paper/2018/file/4c7a167bb329bd92580a99ce422d6fa6-Paper.pdf} {e-snli: Natural language inference with natural language explanations}.
\newblock In \emph{Advances in Neural Information Processing Systems}, volume~31. Curran Associates, Inc.

\bibitem[{Chalkidis et~al.(2022)Chalkidis, Jana, Hartung, Bommarito, Androutsopoulos, Katz, and Aletras}]{chalkidis-etal-2022-lexglue}
Ilias Chalkidis, Abhik Jana, Dirk Hartung, Michael Bommarito, Ion Androutsopoulos, Daniel Katz, and Nikolaos Aletras. 2022.
\newblock \href {https://doi.org/10.18653/v1/2022.acl-long.297} {{L}ex{GLUE}: A benchmark dataset for legal language understanding in {E}nglish}.
\newblock In \emph{Proceedings of the 60th Annual Meeting of the Association for Computational Linguistics (Volume 1: Long Papers)}, pages 4310--4330, Dublin, Ireland. Association for Computational Linguistics.

\bibitem[{Chen et~al.(2023)Chen, Zaharia, and Zou}]{chen2023chatgpts}
Lingjiao Chen, Matei Zaharia, and James Zou. 2023.
\newblock \href {http://arxiv.org/abs/2307.09009} {How is chatgpt's behavior changing over time?}

\bibitem[{Devlin et~al.(2019)Devlin, Chang, Lee, and Toutanova}]{devlin-etal-2019-bert}
Jacob Devlin, Ming-Wei Chang, Kenton Lee, and Kristina Toutanova. 2019.
\newblock \href {https://doi.org/10.18653/v1/N19-1423} {{BERT}: Pre-training of deep bidirectional transformers for language understanding}.
\newblock In \emph{Proceedings of the 2019 Conference of the North {A}merican Chapter of the Association for Computational Linguistics: Human Language Technologies, Volume 1 (Long and Short Papers)}, pages 4171--4186, Minneapolis, Minnesota. Association for Computational Linguistics.

\bibitem[{Ding and Riloff(2016)}]{10.5555/3016100.3016313}
Haibo Ding and Ellen Riloff. 2016.
\newblock Acquiring knowledge of affective events from blogs using label propagation.
\newblock In \emph{Proceedings of the Thirtieth AAAI Conference on Artificial Intelligence}, AAAI'16, page 2935–2942. AAAI Press.

\bibitem[{Dong et~al.(2023)Dong, Xiong, Goyal, Zhang, Chow, Pan, Diao, Zhang, Shum, and Zhang}]{dong2023raft}
Hanze Dong, Wei Xiong, Deepanshu Goyal, Yihan Zhang, Winnie Chow, Rui Pan, Shizhe Diao, Jipeng Zhang, Kashun Shum, and Tong Zhang. 2023.
\newblock \href {http://arxiv.org/abs/2304.06767} {Raft: Reward ranked finetuning for generative foundation model alignment}.

\bibitem[{Emelin et~al.(2021)Emelin, Le~Bras, Hwang, Forbes, and Choi}]{emelin2020moral}
Denis Emelin, Ronan Le~Bras, Jena~D. Hwang, Maxwell Forbes, and Yejin Choi. 2021.
\newblock \href {https://doi.org/10.18653/v1/2021.emnlp-main.54} {Moral stories: Situated reasoning about norms, intents, actions, and their consequences}.
\newblock In \emph{Proceedings of the 2021 Conference on Empirical Methods in Natural Language Processing}, pages 698--718, Online and Punta Cana, Dominican Republic. Association for Computational Linguistics.

\bibitem[{Forbes et~al.(2020)Forbes, Hwang, Shwartz, Sap, and Choi}]{forbes2020social}
Maxwell Forbes, Jena~D. Hwang, Vered Shwartz, Maarten Sap, and Yejin Choi. 2020.
\newblock \href {https://doi.org/10.18653/v1/2020.emnlp-main.48} {Social chemistry 101: Learning to reason about social and moral norms}.
\newblock In \emph{Proceedings of the 2020 Conference on Empirical Methods in Natural Language Processing (EMNLP)}, pages 653--670, Online. Association for Computational Linguistics.

\bibitem[{Fulgoni et~al.(2016)Fulgoni, Carpenter, Ungar, and Preo{\c{t}}iuc-Pietro}]{fulgoni-etal-2016-empirical}
Dean Fulgoni, Jordan Carpenter, Lyle Ungar, and Daniel Preo{\c{t}}iuc-Pietro. 2016.
\newblock \href {https://aclanthology.org/L16-1591} {An empirical exploration of moral foundations theory in partisan news sources}.
\newblock In \emph{Proceedings of the Tenth International Conference on Language Resources and Evaluation ({LREC}'16)}, pages 3730--3736, Portoro{\v{z}}, Slovenia. European Language Resources Association (ELRA).

\bibitem[{Gangi~Reddy et~al.(2022)Gangi~Reddy, Chinthakindi, Fung, Small, and Ji}]{gangi-reddy-etal-2022-zero}
Revanth Gangi~Reddy, Sai~Chetan Chinthakindi, Yi~R. Fung, Kevin Small, and Heng Ji. 2022.
\newblock \href {https://aclanthology.org/2022.coling-1.603} {A zero-shot claim detection framework using question answering}.
\newblock In \emph{Proceedings of the 29th International Conference on Computational Linguistics}, pages 6927--6933, Gyeongju, Republic of Korea. International Committee on Computational Linguistics.

\bibitem[{Graham et~al.(2009)Graham, Haidt, and Nosek}]{Graham2009MoralFound}
Jesse Graham, Jonathan Haidt, and Brian Nosek. 2009.
\newblock Liberals and conservatives rely on different sets of moral foundations.
\newblock In \emph{Journal of Personality and Social Psychology}, pages 1029--1046. Brill.

\bibitem[{Grice(1975)}]{grice1975logic}
Herbert~P Grice. 1975.
\newblock Logic and conversation.
\newblock In \emph{Speech acts}, pages 41--58. Brill.

\bibitem[{Han et~al.(2023)Han, Xu, Li, Fung, Sun, Abdelzaher, and Ji}]{LMSwitch2023}
Chi Han, Jialiang Xu, Manling Li, Yi~R. Fung, Chenkai Sun, Tarek Abdelzaher, and Heng Ji. 2023.
\newblock Lm-switch: Lightweight language model conditioning in word embedding space.
\newblock In \emph{arxiv}.

\bibitem[{Huang et~al.(2023)Huang, Chan, and Ji}]{factualerrorcorrection2023}
Kung-Hsiang Huang, Hou~Pong Chan, and Heng Ji. 2023.
\newblock Zero-shot faithful factual error correction.
\newblock In \emph{Proc. The 61st Annual Meeting of the Association for Computational Linguistics (ACL2023)}.

\bibitem[{Huang et~al.(2022)Huang, Zhai, and Ji}]{Huang2022}
Kung-Hsiang Huang, ChengXiang Zhai, and Heng Ji. 2022.
\newblock Improving cross-lingual fact checking with cross-lingual retrieval.
\newblock In \emph{Proc. The 29th International Conference on Computational Linguistics (COLING2022)}.

\bibitem[{Jiang et~al.(2021{\natexlab{a}})Jiang, Hwang, Bhagavatula, Bras, Forbes, Borchardt, Liang, Etzioni, Sap, and Choi}]{jiang2021delphi}
Liwei Jiang, Jena~D Hwang, Chandra Bhagavatula, Ronan~Le Bras, Maxwell Forbes, Jon Borchardt, Jenny Liang, Oren Etzioni, Maarten Sap, and Yejin Choi. 2021{\natexlab{a}}.
\newblock Delphi: Towards machine ethics and norms.
\newblock \emph{arXiv preprint arXiv:2110.07574}.

\bibitem[{Jiang et~al.(2021{\natexlab{b}})Jiang, Araki, Ding, and Neubig}]{jiang-etal-2021-know}
Zhengbao Jiang, Jun Araki, Haibo Ding, and Graham Neubig. 2021{\natexlab{b}}.
\newblock \href {https://doi.org/10.1162/tacl_a_00407} {How can we know when language models know? on the calibration of language models for question answering}.
\newblock \emph{Transactions of the Association for Computational Linguistics}, 9:962--977.

\bibitem[{Kallia(2004)}]{kallia2004linguistic}
Alexandra Kallia. 2004.
\newblock Linguistic politeness: The implicature approach.
\newblock \emph{Multilingua}, 23.

\bibitem[{Le~Scao and Rush(2021)}]{le-scao-rush-2021-many}
Teven Le~Scao and Alexander Rush. 2021.
\newblock \href {https://doi.org/10.18653/v1/2021.naacl-main.208} {How many data points is a prompt worth?}
\newblock In \emph{Proceedings of the 2021 Conference of the North American Chapter of the Association for Computational Linguistics: Human Language Technologies}, pages 2627--2636, Online. Association for Computational Linguistics.

\bibitem[{Lei et~al.(2018)Lei, Yu, Bansal, and Berg}]{lei-etal-2018-tvqa}
Jie Lei, Licheng Yu, Mohit Bansal, and Tamara Berg. 2018.
\newblock \href {https://doi.org/10.18653/v1/D18-1167} {{TVQA}: Localized, compositional video question answering}.
\newblock In \emph{Proceedings of the 2018 Conference on Empirical Methods in Natural Language Processing}, pages 1369--1379, Brussels, Belgium. Association for Computational Linguistics.

\bibitem[{Lewis et~al.(2020)Lewis, Liu, Goyal, Ghazvininejad, Mohamed, Levy, Stoyanov, and Zettlemoyer}]{lewis-etal-2020-bart}
Mike Lewis, Yinhan Liu, Naman Goyal, Marjan Ghazvininejad, Abdelrahman Mohamed, Omer Levy, Veselin Stoyanov, and Luke Zettlemoyer. 2020.
\newblock \href {https://doi.org/10.18653/v1/2020.acl-main.703} {{BART}: Denoising sequence-to-sequence pre-training for natural language generation, translation, and comprehension}.
\newblock In \emph{Proceedings of the 58th Annual Meeting of the Association for Computational Linguistics}, pages 7871--7880, Online. Association for Computational Linguistics.

\bibitem[{Li et~al.(2023{\natexlab{a}})Li, Han, Yu, Edwards, Li, Wang, Fung, Yu, Tetreault, and Hovy}]{Li2023NewPlayground}
Sha Li, Chi Han, Pengfei Yu, Carl Edwards, Manling Li, Xingyao Wang, Yi~R. Fung, Charles Yu, Joel~R. Tetreault, and Heng Hovy, Eduard H;~Ji. 2023{\natexlab{a}}.
\newblock Defining a new nlp playground.
\newblock \emph{ACL Findings}.

\bibitem[{Li et~al.(2023{\natexlab{b}})Li, Zhao, Li, Ji, Callison-Burch, and Han}]{hierarchicalschema2023}
Sha Li, Ruining Zhao, Manling Li, Heng Ji, Chris Callison-Burch, and Jiawei Han. 2023{\natexlab{b}}.
\newblock Open-domain hierarchical event schema induction by incremental prompting and verification.
\newblock In \emph{Proc. The 61st Annual Meeting of the Association for Computational Linguistics (ACL2023)}.

\bibitem[{Lin et~al.(2018)Lin, Hoover, Portillo-Wightman, Park, Dehghani, and Ji}]{Lin2018c}
Ying Lin, Joe Hoover, Gwenyth Portillo-Wightman, Christina Park, Morteza Dehghani, and Heng Ji. 2018.
\newblock Acquiring background knowledge to improve moral value prediction.
\newblock In \emph{Proc. The 2018 IEEE/ACM International Conference on Advances in Social Networks Analysis and Mining (ASONAM2018)}.

\bibitem[{Lourie et~al.(2021)Lourie, Le~Bras, and Choi}]{lourie2021scruples}
Nicholas Lourie, Ronan Le~Bras, and Yejin Choi. 2021.
\newblock Scruples: A corpus of community ethical judgments on 32,000 real-life anecdotes.
\newblock In \emph{Proceedings of the AAAI Conference on Artificial Intelligence}, volume 35(15), pages 13470--13479.

\bibitem[{Mathew et~al.(2021)Mathew, Saha, Yimam, Biemann, Goyal, and Mukherjee}]{mathew2021hatexplain}
Binny Mathew, Punyajoy Saha, Seid~Muhie Yimam, Chris Biemann, Pawan Goyal, and Animesh Mukherjee. 2021.
\newblock Hatexplain: A benchmark dataset for explainable hate speech detection.
\newblock In \emph{Proceedings of the AAAI Conference on Artificial Intelligence}, volume~35, pages 14867--14875.

\bibitem[{Mooijman et~al.(2018)Mooijman, Hoover, Lin, Ji, and Dehghani}]{Mooijman2018}
Marlon Mooijman, Joe Hoover, Ying Lin, Heng Ji, and Morteza Dehghani. 2018.
\newblock Moralization in social networks and the emergence of violent protests.
\newblock \emph{Nature Human Behavior [\textbf{June 2018 Cover}]}.

\bibitem[{Omrani et~al.(2023)Omrani, Ziabari, Yu, Golazizian, Kennedy, Atari, Ji, and Dehghani}]{debiasing2023b}
Ali Omrani, Alireza~Salkhordeh Ziabari, Charles Yu, Preni Golazizian, Brendan Kennedy, Mohammad Atari, Heng Ji, and Morteza Dehghani. 2023.
\newblock Social-group-agnostic bias mitigation via the stereotype content model.
\newblock In \emph{Proc. The 61st Annual Meeting of the Association for Computational Linguistics (ACL2023)}.

\bibitem[{Ouyang et~al.(2022)Ouyang, Wu, Jiang, Almeida, Wainwright, Mishkin, Zhang, Agarwal, Slama, Ray et~al.}]{ouyang2022training}
Long Ouyang, Jeffrey Wu, Xu~Jiang, Diogo Almeida, Carroll Wainwright, Pamela Mishkin, Chong Zhang, Sandhini Agarwal, Katarina Slama, Alex Ray, et~al. 2022.
\newblock Training language models to follow instructions with human feedback.
\newblock \emph{Advances in Neural Information Processing Systems}, 35:27730--27744.

\bibitem[{Petroni et~al.(2019)Petroni, Rockt{\"a}schel, Riedel, Lewis, Bakhtin, Wu, and Miller}]{petroni-etal-2019-language}
Fabio Petroni, Tim Rockt{\"a}schel, Sebastian Riedel, Patrick Lewis, Anton Bakhtin, Yuxiang Wu, and Alexander Miller. 2019.
\newblock \href {https://doi.org/10.18653/v1/D19-1250} {Language models as knowledge bases?}
\newblock In \emph{Proceedings of the 2019 Conference on Empirical Methods in Natural Language Processing and the 9th International Joint Conference on Natural Language Processing (EMNLP-IJCNLP)}, pages 2463--2473, Hong Kong, China. Association for Computational Linguistics.

\bibitem[{Posner(2009)}]{eric2009normlaw}
Eric Posner. 2009.
\newblock \emph{Law and social norms}, chapter 1-4. Havard University Press.

\bibitem[{Raffel et~al.(2020)Raffel, Shazeer, Roberts, Lee, Narang, Matena, Zhou, Li, Liu et~al.}]{raffel2020exploring}
Colin Raffel, Noam Shazeer, Adam Roberts, Katherine Lee, Sharan Narang, Michael Matena, Yanqi Zhou, Wei Li, Peter~J Liu, et~al. 2020.
\newblock Exploring the limits of transfer learning with a unified text-to-text transformer.
\newblock \emph{J. Mach. Learn. Res.}, 21(140):1--67.

\bibitem[{Rashkin et~al.(2018)Rashkin, Sap, Allaway, Smith, and Choi}]{rashkin-etal-2018-event2mind}
Hannah Rashkin, Maarten Sap, Emily Allaway, Noah~A. Smith, and Yejin Choi. 2018.
\newblock \href {https://doi.org/10.18653/v1/P18-1043} {{E}vent2{M}ind: Commonsense inference on events, intents, and reactions}.
\newblock In \emph{Proceedings of the 56th Annual Meeting of the Association for Computational Linguistics (Volume 1: Long Papers)}, pages 463--473, Melbourne, Australia. Association for Computational Linguistics.

\bibitem[{Reddy et~al.(2023)Reddy, Fung, Zeng, Li, Wang, Sullivan, and Ji}]{reddy2023smartbook}
Revanth~Gangi Reddy, Yi~R. Fung, Qi~Zeng, Manling Li, Ziqi Wang, Paul Sullivan, and Heng Ji. 2023.
\newblock \href {http://arxiv.org/abs/2303.14337} {Smartbook: Ai-assisted situation report generation}.

\bibitem[{Roberts(2016)}]{roberts2016commercial}
Sarah~T Roberts. 2016.
\newblock Commercial content moderation: Digital laborers' dirty work.
\newblock In \emph{The Intersectional Internet: Race, Sex, Class and Culture Online}. Peter Lang Publishing.

\bibitem[{Sanh et~al.(2022)Sanh, Webson, Raffel, Bach, Sutawika, Alyafeai, Chaffin, Stiegler, Raja, Dey, Bari, Xu, Thakker, Sharma, Szczechla, Kim, Chhablani, Nayak, Datta, Chang, Jiang, Wang, Manica, Shen, Yong, Pandey, Bawden, Wang, Neeraj, Rozen, Sharma, Santilli, Fevry, Fries, Teehan, Scao, Biderman, Gao, Wolf, and Rush}]{sanh2022multitask}
Victor Sanh, Albert Webson, Colin Raffel, Stephen Bach, Lintang Sutawika, Zaid Alyafeai, Antoine Chaffin, Arnaud Stiegler, Arun Raja, Manan Dey, M~Saiful Bari, Canwen Xu, Urmish Thakker, Shanya~Sharma Sharma, Eliza Szczechla, Taewoon Kim, Gunjan Chhablani, Nihal Nayak, Debajyoti Datta, Jonathan Chang, Mike Tian-Jian Jiang, Han Wang, Matteo Manica, Sheng Shen, Zheng~Xin Yong, Harshit Pandey, Rachel Bawden, Thomas Wang, Trishala Neeraj, Jos Rozen, Abheesht Sharma, Andrea Santilli, Thibault Fevry, Jason~Alan Fries, Ryan Teehan, Teven~Le Scao, Stella Biderman, Leo Gao, Thomas Wolf, and Alexander~M Rush. 2022.
\newblock \href {https://openreview.net/forum?id=9Vrb9D0WI4} {Multitask prompted training enables zero-shot task generalization}.
\newblock In \emph{International Conference on Learning Representations}.

\bibitem[{Sap et~al.(2019{\natexlab{a}})Sap, Gabriel, Qin, Jurafsky, Smith, and Choi}]{sap2019social}
Maarten Sap, Saadia Gabriel, Lianhui Qin, Dan Jurafsky, Noah~A Smith, and Yejin Choi. 2019{\natexlab{a}}.
\newblock Social bias frames: Reasoning about social and power implications of language.
\newblock \emph{arXiv preprint arXiv:1911.03891}.

\bibitem[{Sap et~al.(2019{\natexlab{b}})Sap, LeBras, Allaway, Bhagavatula, Lourie, Rashkin, Roof, Smith, and Choi}]{sap2019atomic}
Maarten Sap, Ronan LeBras, Emily Allaway, Chandra Bhagavatula, Nicholas Lourie, Hannah Rashkin, Brendan Roof, Noah~A Smith, and Yejin Choi. 2019{\natexlab{b}}.
\newblock \href {https://aaai.org/ojs/index.php/AAAI/article/view/4160} {Atomic: An atlas of machine commonsense for if-then reasoning}.
\newblock In \emph{AAAI}.

\bibitem[{Schwartz et~al.(2012)}]{schwartz2012overview}
Shalom~H Schwartz et~al. 2012.
\newblock An overview of the schwartz theory of basic values.
\newblock \emph{Online readings in Psychology and Culture}, 2(1):2307--0919.

\bibitem[{Sheehan(2018)}]{sheehan2018crowdsourcing}
Kim~Bartel Sheehan. 2018.
\newblock Crowdsourcing research: data collection with amazon’s mechanical turk.
\newblock \emph{Communication Monographs}, 85(1):140--156.

\bibitem[{Solaiman and Dennison(2021)}]{solaiman2021process}
Irene Solaiman and Christy Dennison. 2021.
\newblock Process for adapting language models to society (palms) with values-targeted datasets.
\newblock \emph{Advances in Neural Information Processing Systems}, 34:5861--5873.

\bibitem[{Song et~al.(2014)Song, Strassel, Lee, Walker, Wright, Garland, Fore, Gainor, Cabe, Thomas et~al.}]{song2014collecting}
Zhiyi Song, Stephanie~M Strassel, Haejoong Lee, Kevin Walker, Jonathan Wright, Jennifer Garland, Dana Fore, Brian Gainor, Preston Cabe, Thomas Thomas, et~al. 2014.
\newblock Collecting natural sms and chat conversations in multiple languages: The bolt phase 2 corpus.
\newblock In \emph{LREC}, pages 1699--1704. Citeseer.

\bibitem[{Sun et~al.(2023)Sun, Li, Fung, Chan, Abdelzaher, Zhai, and Ji}]{sun2023}
Chenkai Sun, Jinning Li, Yi~Fung, Hou~P. Chan, Tarek Abdelzaher, Chengxiang Zhai, and Heng Ji. 2023.
\newblock Decoding the silent majority: Inducing belief augmented social graph with large language model for response forecasting.
\newblock \emph{The 2023 Conference on Empirical Methods in Natural Language Processing (EMNLP)}.

\bibitem[{Talat et~al.(2021)Talat, Blix, Valvoda, Ganesh, Cotterell, and Williams}]{talat2021word}
Zeerak Talat, Hagen Blix, Josef Valvoda, Maya~Indira Ganesh, Ryan Cotterell, and Adina Williams. 2021.
\newblock A word on machine ethics: A response to jiang et al.(2021).
\newblock \emph{arXiv preprint arXiv:2111.04158}.

\bibitem[{Van~Hee et~al.(2015)Van~Hee, Lefever, Verhoeven, Mennes, Desmet, De~Pauw, Daelemans, and Hoste}]{van-hee-etal-2015-detection}
Cynthia Van~Hee, Els Lefever, Ben Verhoeven, Julie Mennes, Bart Desmet, Guy De~Pauw, Walter Daelemans, and Veronique Hoste. 2015.
\newblock \href {https://aclanthology.org/R15-1086} {Detection and fine-grained classification of cyberbullying events}.
\newblock In \emph{Proceedings of the International Conference Recent Advances in Natural Language Processing}, pages 672--680, Hissar, Bulgaria. INCOMA Ltd. Shoumen, BULGARIA.

\bibitem[{Vu et~al.(2014)Vu, Neubig, Sakti, Toda, and Nakamura}]{vu-etal-2014-acquiring}
Hoa~Trong Vu, Graham Neubig, Sakriani Sakti, Tomoki Toda, and Satoshi Nakamura. 2014.
\newblock \href {https://doi.org/10.3115/v1/E14-4025} {Acquiring a dictionary of emotion-provoking events}.
\newblock In \emph{Proceedings of the 14th Conference of the {E}uropean Chapter of the Association for Computational Linguistics, volume 2: Short Papers}, pages 128--132, Gothenburg, Sweden. Association for Computational Linguistics.

\bibitem[{Welleck et~al.(2018)Welleck, Weston, Szlam, and Cho}]{welleck2018dialogue}
Sean Welleck, Jason Weston, Arthur Szlam, and Kyunghyun Cho. 2018.
\newblock Dialogue natural language inference.
\newblock \emph{arXiv preprint arXiv:1811.00671}.

\bibitem[{Wen et~al.(2021)Wen, Lin, Lai, Pan, Li, Lin, Zhou, Li, Wang, Zhang, Yu, Dong, Wang, Fung, Mishra, Lyu, Sur{\'\i}s, Chen, Brown, Palmer, Callison-Burch, Vondrick, Han, Roth, Chang, and Ji}]{wen-etal-2021-resin}
Haoyang Wen, Ying Lin, Tuan Lai, Xiaoman Pan, Sha Li, Xudong Lin, Ben Zhou, Manling Li, Haoyu Wang, Hongming Zhang, Xiaodong Yu, Alexander Dong, Zhenhailong Wang, Yi~Fung, Piyush Mishra, Qing Lyu, D{\'\i}dac Sur{\'\i}s, Brian Chen, Susan~Windisch Brown, Martha Palmer, Chris Callison-Burch, Carl Vondrick, Jiawei Han, Dan Roth, Shih-Fu Chang, and Heng Ji. 2021.
\newblock \href {https://doi.org/10.18653/v1/2021.naacl-demos.16} {{RESIN}: A dockerized schema-guided cross-document cross-lingual cross-media information extraction and event tracking system}.
\newblock In \emph{Proceedings of the 2021 Conference of the North American Chapter of the Association for Computational Linguistics: Human Language Technologies: Demonstrations}, pages 133--143, Online. Association for Computational Linguistics.

\bibitem[{Wiegreffe et~al.(2021)Wiegreffe, Hessel, Swayamdipta, Riedl, and Choi}]{wiegreffe2021reframing}
Sarah Wiegreffe, Jack Hessel, Swabha Swayamdipta, Mark Riedl, and Yejin Choi. 2021.
\newblock Reframing human-ai collaboration for generating free-text explanations.
\newblock \emph{arXiv preprint arXiv:2112.08674}.

\bibitem[{Williams et~al.(2018)Williams, Nangia, and Bowman}]{N18-1101}
Adina Williams, Nikita Nangia, and Samuel Bowman. 2018.
\newblock \href {http://aclweb.org/anthology/N18-1101} {A broad-coverage challenge corpus for sentence understanding through inference}.
\newblock In \emph{Proceedings of the 2018 Conference of the North American Chapter of the Association for Computational Linguistics: Human Language Technologies, Volume 1 (Long Papers)}, pages 1112--1122. Association for Computational Linguistics.

\bibitem[{Yang et~al.(2023)Yang, Yu, Fung, Li, and Ji}]{Adept2023}
Ke~Yang, Charles Yu, Yi~Fung, Manling Li, and Heng Ji. 2023.
\newblock Adept: A debiasing prompt framework.
\newblock In \emph{Proc. Thirty-Seventh AAAI Conference on Artificial Intelligence (AAAI2023)}.

\bibitem[{Ye and Durrett(2022)}]{NEURIPS2022_c4025018}
Xi~Ye and Greg Durrett. 2022.
\newblock \href {https://proceedings.neurips.cc/paper_files/paper/2022/file/c402501846f9fe03e2cac015b3f0e6b1-Paper-Conference.pdf} {The unreliability of explanations in few-shot prompting for textual reasoning}.
\newblock In \emph{Advances in Neural Information Processing Systems}, volume~35, pages 30378--30392. Curran Associates, Inc.

\bibitem[{Yu et~al.(2023)Yu, Jeoung, Kasi, Yu, and Ji}]{debiasinggradient2023}
Charles Yu, Sullam Jeoung, Anish Kasi, Pengfei Yu, and Heng Ji. 2023.
\newblock Unlearning bias in language models by partitioning gradients.
\newblock In \emph{Proc. The 61st Annual Meeting of the Association for Computational Linguistics (ACL2023) Findings}.

\bibitem[{Yu et~al.(2020)Yu, Xu, Meng, Zhu, Ma, Wu, Zou, and Yang}]{yu-etal-2020-ch}
Wenmeng Yu, Hua Xu, Fanyang Meng, Yilin Zhu, Yixiao Ma, Jiele Wu, Jiyun Zou, and Kaicheng Yang. 2020.
\newblock \href {https://doi.org/10.18653/v1/2020.acl-main.343} {{CH}-{SIMS}: A {C}hinese multimodal sentiment analysis dataset with fine-grained annotation of modality}.
\newblock In \emph{Proceedings of the 58th Annual Meeting of the Association for Computational Linguistics}, pages 3718--3727, Online. Association for Computational Linguistics.

\bibitem[{Zadeh et~al.(2018)Zadeh, Liang, Poria, Cambria, and Morency}]{zadeh2018multimodal}
AmirAli~Bagher Zadeh, Paul~Pu Liang, Soujanya Poria, Erik Cambria, and Louis-Philippe Morency. 2018.
\newblock Multimodal language analysis in the wild: Cmu-mosei dataset and interpretable dynamic fusion graph.
\newblock In \emph{Proceedings of the 56th Annual Meeting of the Association for Computational Linguistics (Volume 1: Long Papers)}, pages 2236--2246.

\bibitem[{Ziems et~al.(2022)Ziems, Yu, Wang, Halevy, and Yang}]{ziems2022moral}
Caleb Ziems, Jane Yu, Yi-Chia Wang, Alon Halevy, and Diyi Yang. 2022.
\newblock \href {https://doi.org/10.18653/v1/2022.acl-long.261} {The moral integrity corpus: A benchmark for ethical dialogue systems}.
\newblock In \emph{Proceedings of the 60th Annual Meeting of the Association for Computational Linguistics (Volume 1: Long Papers)}, pages 3755--3773, Dublin, Ireland. Association for Computational Linguistics.

\end{thebibliography}
\bibliographystyle{acl_natbib}

\appendix
\section{Appendix}
\subsection{Speaker Profiling and Cultural Indicator Extraction for Norm Discovery with Frames} \label{A.1}
As discussed in Sec \ref{sec:3}, profiling speakers with cultural indicators help guide better structure and culture-specific insights in norm discovery from conversations on-the-fly. The prompting procedure is intuitive with the presentation of the dialogue content and a directed question describing the task, similar to the prompting logic we introduced for conversation-based norm discovery, norm correctness verification, and relevance grounding. The specific prompt template used for cultural indicator extraction is:
\begin{quote}
    \textit{Context:}\\
    $\{X_{s}\}$ \\[4pt]
    \textit{For each person, extract the country- or state- level culture they are affiliated (in adjective form) if the information is available, and skip the person if the information is not available:}
\end{quote}
For instance, the input for a dialogue snippet from "Outsourced" comes with a metadata description of summary from its Wikipedia page, containing text content such as:
\begin{quote}
    \textit{"Todd Anderson (Josh Hamilton), a salesman for a Seattle novelty products company, learns he has to travel to India when his department is outsourced. Todd is not happy but when his boss Dave informs him that quitting would mean losing his stock options, he goes to train his Indian replacement Puro (Asif Basra)."}, 
\end{quote}
which we feed as value into $\{X_{b}\}$ in the prompt template. The output of this cultural indicator extraction is "Todd Anderson: American" and "Puro: Indian" in separate lines, which follows a consistent textual pattern that is easy to parse and is then plugged into the \textit{dvr($\cdot$)}-\textsc{NormSage} input for norm discovery with framing.

\subsection{Norm Discovery Intrinsic Evaluation} 
\subsubsection{Assessment Metric Guidelines}\label{A:2.1}
We provide definitions of each intrinsic metric to the crowdsourced annotation workers. 
\begin{itemize}[nosep,leftmargin=*]
   \itemsep0em 
    \item \textit{\textbf{Relevance}}: is the norm inspired from the situation (lower bound on norm applicability). 
    \item \textit{\textbf{Well-Formedness}}: how well is the norm structured -- is the norm self-contained, and does it include \textit{both} a judgment of acceptability or occurrence, \textit{and} an action or societal/cultural phenomena that is assessed. 
    \item \textit{\textbf{Correctness}}: to the best of their knowledge, do people agree that the described norm holds true? 
    \item \textit{\textbf{Insightfulness}}: does the norm convey enlightening understanding about what's considered acceptable and standard in the society that pertain to the conversation scenario.
    \item \textit{\textbf{Relatableness}}: how well does the norm balance vagueness against specificity, so that it can generalize across multiple situations without being too specific.
\end{itemize}

\noindent We also provide good and bad examples of norms discovered, according to each of the metric dimensions. Consider a dialogue instance of \textit{"I think we should divide the project tasks based on everyone's expertise and skills. It will ensure better efficiency and quality"}. We list good and bad examples of social norms with respect to the dialogue scenario as follows.
\begin{itemize}[nosep,leftmargin=*]
\itemsep0em 
    \item Relevant social norm: {\small "It's important to play to people's strengths and utilize their individual knowledge to deliver the best results."} \\
    Irrelevant social norm: {\small"It is good to earn money and retire early."}
    \item Well-formed social norm: {\small"It is good to utilize play to people's strength and deliver the best results".} \\
    Poor-formed social norm: {\small"Deliver the best results."}
    \item Correct social norm: {\small"Assigning tasks based on individual expertise and skills leads to better project outcomes".} \\
    Incorrect social norm: {\small"Assigning tasks based on the alphabetical order of employees' last names leads to better project outcomes".}
    \item Insightful social norm: {\small"Assigning tasks based on individual expertise not only maximizes efficiency and quality but also fosters a sense of ownership, motivation, and collaboration among team members."} \\
    Non-insightful social norm: {\small"It is nice to deliver good results."}
    \item Relatable social norm: {\small"Assigning tasks based on each team member's proficiency in relevant software tools (such as Excel for data analysis and presentation task) can lead to better project outcomes."} \\
    Non-relatable social norm: {\small"Assigning tasks based on each team member's proficiency in the specific Excel software tool leads to better project outcomes".}
\end{itemize} 

We ask annotators to rate social norms discovered from conversations on a 1-5 Likert scale, in which the spectrum of scoring guidelines is illustrated as:
\begin{figure}[h]
    \centering
    \includegraphics[trim=0 0 0 0cm, width=0.47\textwidth]{figures/Likert.pdf}
    \label{fig:likert}
\end{figure}

\label{sec:appendix}

\end{document}